\documentclass[10pt, a4paper]{article}
\usepackage{lrec-coling2024} 

\PassOptionsToPackage{backref=page}{hyperref}

\newif\ifarxiv
\arxivfalse 

\usepackage{twemojis}
\usepackage{times}
\usepackage{latexsym}
\usepackage{booktabs}

\usepackage[T1]{fontenc}

\usepackage[utf8]{inputenc}

\usepackage{microtype}

\usepackage{graphicx}
\usepackage{enumitem,kantlipsum}

\usepackage{amsmath,amsfonts,bm}

\def\eqref#1{equation~\ref{#1}}

\def\1{\bm{1}}

\DeclareMathAlphabet{\mathsfit}{\encodingdefault}{\sfdefault}{m}{sl}
\SetMathAlphabet{\mathsfit}{bold}{\encodingdefault}{\sfdefault}{bx}{n}

\usepackage{hyperref}
\usepackage{url}
\graphicspath{{img/}}

\usepackage{cleveref}
\crefname{figure}{Figure}{Figures}
\crefname{table}{Table}{Tables}
\crefname{appendix}{Appendix}{Appendices}
\crefname{section}{Section}{Sections}
\crefname{equation}{Eq.}{Eqs.}
\crefname{enumi}{}{} 

\usepackage{xspace}

\usepackage{tikz}

\newcommand{\shrug}[1][]{%
\begin{tikzpicture}[baseline,x=0.8\ht\strutbox,y=0.8\ht\strutbox,line width=0.125ex,#1]
\def\arm{(-2.5,0.95) to (-2,0.95) (-1.9,1) to (-1.5,0) (-1.35,0) to (-0.8,0)};
\draw \arm;
\draw[xscale=-1] \arm;
\def\headpart{(0.6,0) arc[start angle=-40, end angle=40,x radius=0.6,y radius=0.8]};
\draw \headpart;
\draw[xscale=-1] \headpart;
\def\eye{(-0.075,0.15) .. controls (0.02,0) .. (0.075,-0.15)};
\draw[shift={(-0.3,0.8)}] \eye;
\draw[shift={(0,0.85)}] \eye;
\draw (-0.1,0.2) to [out=15,in=-100] (0.4,0.95); 
\end{tikzpicture}}

\name{Oana Ignat$^*$, Zhijing Jin$^*$, Artem Abzaliev, Laura Biester, Santiago Castro, \\
{\bf Naihao Deng, Xinyi Gao, Aylin Gunal, Jacky He, Ashkan Kazemi, Muhammad Khalifa, }\\
{\bf Namho Koh, Andrew Lee,  Siyang Liu, Do June Min, Shinka Mori, Joan Nwatu,}\\
{\bf Veronica Perez-Rosas, Siqi Shen, Zekun Wang, Winston Wu, Rada Mihalcea}}

\address{University of Michigan \\ {\sc Language and Information Technologies (LIT)} \\
\tt \{oignat,jinzhi,mihalcea\}@umich.edu}

\title{
Has It All Been Solved? Open NLP Research Questions \\Not Solved by Large Language Models
}

\abstract{
Recent progress in large language models (LLMs) has enabled the deployment of many generative NLP applications. At the same time, it has also led to a misleading public discourse that ``it's all been solved.'' Not surprisingly, this has, in turn, made many NLP researchers -- especially those at the beginning of their careers -- worry about what NLP research area they should focus on.  \textit{Has it all been solved, or what remaining questions can we work on regardless of LLMs?}  To address this question, this paper compiles NLP research directions rich for exploration. We identify fourteen different research areas encompassing 45 research directions that require new research and are not directly solvable by LLMs. While we identify many research areas, many others exist; we do not cover areas currently addressed by LLMs, but where LLMs lag behind in performance or those focused on LLM development.
We welcome suggestions for other research directions to include: \url{https://bit.ly/nlp-era-llm}
\\ \newline } 

\begin{document}

\maketitleabstract

\def\thefootnote{{$*$}}\footnotetext{Oana Ignat and Zhijing Jin contributed equally to the manuscript.}

\section{Background}

Language models represent one of the fundamental building blocks in NLP, with their roots traced back to 1948 when Claude Shannon introduced Markov chains to model sequences of letters in English text \citep{Shannon1948-iw}. They were then heavily used in connection with early research on statistical machine translation \citep{Brown1988-xb,Wilkes1994-hh} and statistical speech processing \citep{Jelinek1976-jp}. While these models have always been an integral part of broad application categories such as text classification, information retrieval, or text generation, only in recent years have they found a ``life of their own’’ with widespread use and deployment. 

The impressive advancements we have witnessed in current 
large
language models (LLMs) directly result from those earlier models. They build on the same simple yet groundbreaking idea: given a series of previous words or characters, we can predict what will come next.  The new 
LLMs
benefit from two main developments: (1) the proliferation of Web 2.0 and user-generated data, which has led to a sharp increase in the availability of data; and (2) the growth in computational capabilities through the introduction of GPUs. Together, these developments have facilitated the resurgence of neural networks (or deep learning) and the availability of very large training datasets for these models.  

\begin{figure}
\includegraphics[width=0.48\textwidth]{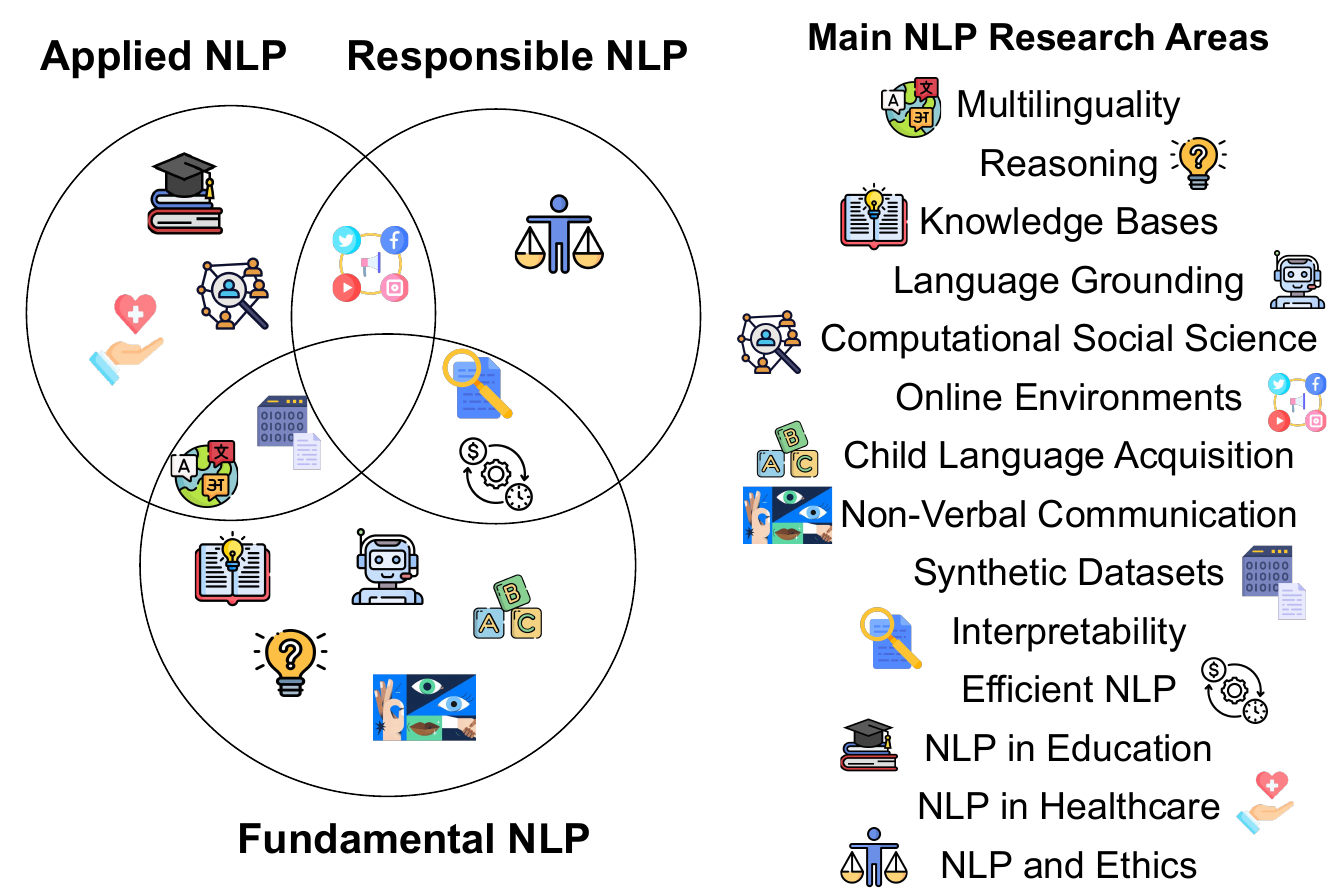}
\caption{Research areas in NLP that are rich for exploration, spanning Fundamental, Responsible, and Applied research.}
\label{fig:venn-grouping}
\end{figure}

Current LLMs have output quality comparable to human performance, with the added benefit of integrating information from enormous data sources, far surpassing what one individual can accumulate in their lifetime. The number of applications that benefit from using LLMs is continuously growing, with many cases where the LLMs are used to replace entire complex pipelines. LLMs becoming ``lucrative’’ has led to a surge in industry interest and funding, alongside a sharp increase in the number of research publications on LLMs. 

While these advances in LLMs are very real and truly exciting, and give hope for many new  generative language applications, LLMs have also ``sucked the air out of the room.'' A recent funding call from DARPA has completely replaced the term NLP with LLM: in their listing of experts sought for the program, we see the fields of ``Computer Vision'' and ``Machine Learning'' listed alongside ``Large Language Models'' (but not ``Natural Language Processing'').\footnote{\url{https://apply.knowinnovation.com/darpaaiforward/}}  Replacing NLP with LLMs is problematic for two main reasons. First, the space of language insights, methods, and broad applications in NLP is much vaster than what can be accomplished by simply predicting the next word. Second, even if not technologically new, LLMs still represent an exclusionary space because of the amount of data and computation required to train.


This public discourse that often reduces the entire field of NLP to the much smaller space of LLMs is not surprisingly leading to a dilemma for those who have dedicated their careers to advancing NLP research, and especially for junior PhD students who have only recently embarked on the path of becoming NLP researchers. ``\textit{\textbf{What should I work on?}}'' is a question we hear now much more often than before, often as a reaction to the misleading thought that ``it’s been all solved.''

The reality is that there is much more to NLP than just LLMs. 
This paper aims to answer the question: ``What are rich areas of exploration in the field of NLP that could lead to a PhD thesis and cover a space that is not within the purview of LLMs.'' Spoiler alert: there are many such research areas!

\paragraph{About This Document.}
This document reflects the ideas about ``the future of NLP research'' from the members of an academic NLP research lab in the United States. The Language and Information Technologies (\twemoji{fire}LIT) lab at the University of Michigan includes students at various stages in their degree, starting with students who are about to embark on a Ph.D., all the way to students who recently completed a Ph.D. degree. The LIT students come from a wide variety of backgrounds, including China, Iran, Japan, Mexico, Nigeria, Romania, Russia, South Korea, the United States, and Uruguay, reflecting a very diverse set of beliefs, values, and lived experiences. Our research interests cover a wide range of NLP areas, including computational social science, causal reasoning, misinformation detection, healthcare conversation analysis, knowledge-aware generation, commonsense reasoning, cross-cultural models, multimodal question answering, non-verbal communication, visual understanding, and more. 

We provide a list of open research questions that are not solved by LLMs. As showed in \cref{fig:venn-grouping}, we cover three major categories, from fundamental NLP (\cref{sec:fundamental}), to responsible NLP (\cref{sec:responsible}), and applied NLP (\cref{sec:applied}). Spanning across the three categories, we cover 14 open research topics, each with three to four specific research directions. Note that when a research topic touches multiple categories, for convenience, we list them under the major one.

When compiling the ideas in this document, we followed three main guiding principles. First, we aimed to identify areas of NLP research that are rich for exploration, e.g., areas one could write a Ph.D. thesis on. Second, we wanted to highlight research directions that do not have a direct dependency on a paid resource; while the use of existing paid APIs can be fruitful for certain tasks, such as the construction of synthetic datasets, building systems that cannot function without paid APIs is not well aligned with academic core research goals. Third, we targeted research directions that can find solutions with reasonable computational costs achievable with setups more typically available in academic labs. 
Finally, we found inspiration in the ACL list of research areas, from which we selected the ones not in the purview of LLMs.\footnote{We used the ACL 2018 list of areas. 
The mapping of our research areas and the ACL tracks
can be found in the Appendix in \Cref{tab:area_map}.}

Our brainstorming process started with ideas written on sticky notes by all the authors of this document, followed by a ``clustering'' process where we grouped the initial ideas and identified several main themes. These initial themes were then provided to small groups of 2--3 students, who discussed them, expanded or merged some of the themes, and identified several directions worthy of exploration. The final set of themes formed the seed of this document. Each research area has then had multiple passes from multiple students (and Rada) to delineate the background of each theme, the gaps, and the most promising research directions. 
\paragraph{Disclaimer.} The research areas listed in this document are just a few of the areas rich in exploration; many others exist. In particular, we have not listed the numerous research directions where LLMs have been demonstrated to lag in performance \cite{bang2023multitask}, including information extraction, question answering, text summarization, and others. We have also not listed the research directions focused on LLM development, as that is already a major focus in many current research papers, and our goal was to highlight the research directions other than LLM development. We welcome suggestions for other research areas or directions to include: \url{https://bit.ly/nlp-era-llm}
\vspace{-3mm}
\paragraph{Document Organization.} 
We provide a list of open research questions that are not solved by LLMs. As showed in \cref{fig:venn-grouping}, we cover three major categories, from fundamental NLP (\cref{sec:fundamental}), to responsible NLP (\cref{sec:responsible}), and applied NLP (\cref{sec:applied}). Spanning across the three categories, we cover 14 open research topics, each with three to four specific research directions. Note that when a research topic touches multiple categories, for convenience, we list them under the major one.

The following sections provide brief descriptions of fourteen research areas rich in exploration, each with 3--4 research directions. These areas can be  divided into areas that cannot be solved by LLMs for being too data-hungry or for lacking reasoning or grounding abilities (subsections 2.1--2.5, 3.3, 4.3); areas for which we cannot use LLMs because of not having the right data (subsections 2.6, 4.1, 4.2); or areas that could contribute to improving the abilities and quality of LLMs (subsections 3.1, 3.2, 3.4, 4.4). 
When compiling the research directions, we follow several guiding principles. First, we aim to identify areas that are rich for exploration, that one could write a PhD thesis on. Second, we want to highlight research directions that do not directly depend on a paid resource, such as a paid API. Third, we target research directions that can find solutions with reasonable computational costs, achievable in academic labs. Finally, we find inspiration from ACL 2018 list of research areas, from which we select the ones not in the purview of LLMs (15/21 areas). 
The mapping of our research areas and the ACL 2018 tracks
can be found in Appendix \Cref{tab:area_map}.

\section{Fundamental NLP}\label{sec:fundamental}

Fundamental NLP tasks represent a significant subset of NLP research, as illustrated in \cref{fig:flow-chart}.  Among these,
we first consider different ``L''s in ``NLP,'' namely different choices of languages. Although most NLP tasks and datasets use English as a medium, there is a growing trend to extend NLP to more non-English languages (\cref{sec:multi}), child language (\cref{sec:child}), and non-verbal communication (\cref{sec:nonverbal}).
Moreover, there are different ``P''s in ``NLP'' too, where we consider different types of processing tasks on text, such as reasoning (\cref{sec:reason}), knowledge bases (\cref{sec:kb}), and language grounding (\cref{sec:ground}).

In the following, we present the main research directions for each of these research areas.

\subsection
{\includegraphics[scale=0.2]{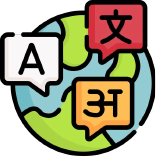} Multilinguality} \label{sec:multi}

\paragraph{Low-resource machine translation.}
Despite the impressive performance of machine translation (MT) on major languages {\citep{hassan2018, mbart2020}, there is a big gap when it comes to low-resource languages. There has been a rise in small benchmarks dedicated to low-resource languages 
\citep{vegi-etal-2022-anvita, reid-etal-2021-afromt, flores101}, but there is also a big need for large training corpora. As many 
low-resource languages do not have a significant web presence, alternative solutions are needed, such as manually curated parallel corpora \citep{zheng-etal-2022-parallel,koto-koto-2020-towards}, 
OCR \citep{Rijhwani2020-sg, Ignat2022-di},
or translation dictionaries using models of word formation \citep{wu-yarowsky-2018-creating,wu-yarowsky-2020-computational}.

\vspace{-3mm}
\paragraph{Multilingual models that work well for all languages.} Although most recent LLMs claim to be multilingual, they do not perform equally well in all languages {\citep{openai2023gpt4, notequal2023, mega2023}}. This inequality stems from the different proportions of text of different languages in the training corpora, 
as well as the annotators with demographics focused in a few countries in the RLHF process to finetune the models \citep{rlhf2022}. 
At present, general-purpose LLMs do not perform as well as models trained specifically for translation; future research can explore incorporating off-the-shelf LLMs into MT systems.


\vspace{-3mm}
\paragraph{Code-switching.} 
Code-switching refers to text involving
expressions in several languages while adhering to the grammatical structure of at least one language. 
Challenges include the large variation of code-switching phenomena, lack of training data, and a large number of
out-of-vocabulary tokens \cite{cetinoglu-etal-2016-challenges}.
Open research directions include synthetic data generation
\cite{Xu2021CanYT,FangWu+2022,Lee2020ModelingCL},
evaluating existing LLMs on code-switched text
across language combinations \cite{aguilar-etal-2020-lince,khanuja-etal-2020-gluecos}, and 
distinguishing highly similar languages, such as dialects of the same parent language \cite{aguilar-etal-2020-lince}.


\subsection{\includegraphics[scale=0.2]{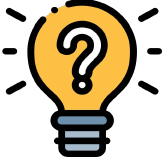} Reasoning}

\label{sec:reason}

\paragraph{Complex reasoning.} 
Complex and multi-step reasoning has proven to be challenging for LLMs. For instance, LLMs still fall short in numerical reasoning \cite{stolfo-etal-2023-causal,miao-etal-2020-diverse}, 
logical reasoning \cite{jin-etal-2022-logical,eisape2023systematic}, grounded reasoning \cite{ignat2021whyact},
and causal inference \cite{jin2023causalbenchmark,jin2023large}, often making obvious mistakes \cite{goel-etal-2021-robustness,jin2020bert}. One reason for that is that the next-word prediction objective can easily encourage the LM to assign a high likelihood to invalid reasoning \cite{khalifa2023grace}. Even fully-supervised training over correct reasoning demonstrations does not solve the issue \cite{uesato2022solving}.  While scaling seems to help, careful prompt engineering is still needed to tease out correct reasoning \cite{wei2022chain,fu2022complexity,zhou2022least,zhang2022automatic}.
To build LLMs that are robust at reasoning, one could explore a variety of directions, such as combining the strengths of neural networks and symbolic AI.
Another growing direction is the integration of LLMs with external reasoning tools, such as calculators, interpreters, database interfaces, or search engines \cite{schick2023toolformer,mialon2023augmented}. 

\vspace{-3.9mm}
\paragraph{Responsible reasoning in social contexts.} 
With an increasing number of applications that use NLP models, it is foreseeable that models will need to make complicated decisions that involve moral reasoning as intermediate steps. For example, when creating a website, there may be moral choices to consider such as catering to certain sub-populations, or overly optimizing for user attention or click-through rates. These decision principles are pervasive in our daily life, across small and large tasks.
We believe there is much to be studied in understanding or improving the ability of AI systems to reason over socially-complicated and morally-charged scenarios given different social contexts and cultural backgrounds \citep{jin2023trolley,hendrycks2021aligning,liu-etal-2021-towards}. 
We foresee that interdisciplinary collaboration with domain experts and policymakers will be needed.
\vspace{-3mm}
\paragraph{Formally defining reasoning and designing a proper evaluation framework.} 
There is a rising need to refine the definition of reasoning, as LLMs start showing an increasing mastery of templated solutions through pattern matching -- when a model memorizes a reasoning pattern, does it count as reasoning or knowledge?
Fundamentally, this leads to questions about what are the domains of intelligence that humans excel at, and how different these are from empirically learning how to do template matching. Beyond redefining reasoning, other open questions include how to test a model's reasoning skills in the face of data contamination, Goodhart's law (a dataset failing to reflect the skill once exploited)~\citep{Goodhart1984ProblemsOM}, and a lack of reliable metrics to evaluate multi-step reasoning \cite{golovneva2022roscoe}. 



\subsection{\includegraphics[scale=0.2]{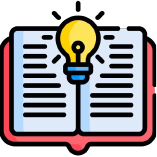} Knowledge Bases}

\label{sec:kb}

\paragraph{Automatic knowledge base construction.} 
Specialized knowledge bases are helpful resources for domain-specific applications.
Successful automatic knowledge base construction can take up-to-date text in free forms \cite{Maedche2000MiningOF}, and adapt an ontology for complex applications, such as tracking medication interactions from articles from PubMed \cite{Xu2020BuildingAP}.\footnote{\url{https://pubmed.ncbi.nlm.nih.gov/}} However, this task faces many challenges,
such as knowledge coverage, factuality of the knowledge, and knowledge linking, which are rich, open areas of research. Specifically, \citet{Wang2023OnTR} shows that ChatGPT performs poorly on out-of-distribution data, such as new medical diagnosis and product review datasets. Also, the training data cutoff limits the coverage of new concepts. 
Specifically concerning factuality, KG completion framed as a text generation task also suffers from hallucination from the LLM  in various tasks~\cite{Ji2022SurveyOH}. 
\vspace{-3mm}
\paragraph{Knowledge-guided NLP.} 
As NLP models become more powerful through exposure to massive pretraining corpora~\citep{hoffmann2022,emergent2022},
researchers start to question whether mere pretraining is sufficient, as models suffer heavily from hallucination ~\citep{hallucination2021survey,hallucinations2022origin}. A rising research question is how to efficiently and effectively interact with external knowledge bases \citep{Zhang2019-qr}, such as through web browsing~\citep{webgpt2021,internetaug2022,schick2023toolformer} and customized knowledge base lookup~\citep{
memory_knowledge2021, augmented2023survey}. 


\vspace{-3mm}
\paragraph{Culture-specific knowledge and common sense.}
Knowledge and common sense in NLP models are usually dominated by a few Western cultures,
and do not account for the vast diversity of the cultural views in the world \citep{arora-etal-2023-probing}. 
The first step is to understand the limitations of NLP models, including LLMs, with respect to their knowledge of different cultural groups \citep{modelsocial2021, crosscluturalnlp2022, arora-etal-2023-probing}. Once these limitations are better understood, a major open research direction is how to acquire and represent the knowledge that encodes these cultural views, as well as how and when to invoke this cultural knowledge.


\subsection{\includegraphics[scale=0.2]{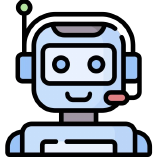} Language Grounding}

\label{sec:ground}
\paragraph{Fusing multiple modalities.} Efficiently and effectively combining different modalities, i.e., audio, video, text, and others, is still an open problem. Different modalities often complement each other, thus potentially reducing the need for billions of data points. However, in some cases, modalities end up competing with each other \cite{yao2022modality}, and thus many uni-modal models outperform multi-modal models \citep{Wang2019-ux, Huang2021-mt}.
\vspace{-3mm}

\paragraph{Grounding for less studied modalities.} Most work on grounding revolves around visual, textual, or acoustic modalities. 
However, less-studied modalities, such as physiological, sensorial, or behavioral, have been found valuable in diverse applications, including measuring driver alertness \citep{Jie2018-pp, Riani2020-to}, detecting depression \citep{Bilalpur2023-ia}, or detecting deceptive behaviors \citep{Abouelenien2016-te}. Current multimodal LLMs are restricted to textual, audio, and visual domains~\cite{Zhang2023VideoLLaMAAI, Lyu2023MacawLLMML}, requiring much effort to integrate the less-studied modalities.
\vspace{-3mm}

\paragraph{Grounding ``in the wild’’ and for diverse domains.} Most research around grounding uses data collected indoors in the lab, or on images and videos of indoor activities from sources such as movies \citep{lei2019tvqa} or online vlogs \citep{Ignat2019IdentifyingVA}. There are fewer studies on outdoor activities in more realistic ``in the wild'' settings 
\citep{Castro2022-el}. Collecting such data poses new challenges related to availability, quality, or distribution, which opens up new research directions. Moreover, applying these models to diverse domains (e.g., robotics, medicine, and education) requires adapting to fewer data points or different types of data, and adding in-domain expertise to understand the problem setup better. As shown in \citet{Yin2023ASO}, the multimodal LLMs are currently not equipped to tackle these challenges.


\subsection{\includegraphics[scale=0.2]{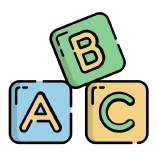} Child Language Acquisition}

\label{sec:child}

\paragraph{Sample-efficient language learning.} 
Child language acquisition is both an important research topic in psycholinguistics \citep{mcneill1970acquisition}, and also a valuable source of inspiration for sample-efficient language learning for NLP \citep{linzen2020can}. By mimicking the learning strategies of children, models can achieve better generalization with limited training data \citep{barak2016comparing}.
Research in this area brings hope to improve the performance of NLP models while reducing the amount of training data required \citep{gulordava2018colorless,warstadt-etal-2023-findings,conll-2023-babylm}.
LLMs require far more data than children to acquire language, and LLMs can be improved in sample efficiency by learning how children acquire language.

\vspace{-3mm}
\paragraph{Language models as biological models for child language acquisition.} 
Since the last century, there has been research using 
neural models as biological models to develop theories of human cognitive behavior
\citep{McCloskey1991}. 
Combining powerful LLMs with psycholinguistic studies, researchers can gain inspiration for various processes in child language acquisition. For instance, insights into word acquisition can be gained by comparing the models' learning curves and children's age of acquisition for different words \citep{chang2021word}. 
Other phenomena, such as phoneme-level acquisition~\citep{christiansen1999connectionist, Martin2023ProbingSS} or intrinsic rewards~\citep{Gibson2019HowES,mu2022improving},  can also be explored by using computational models on existing benchmarks such as WordBank \citep{FRANK2016} or CHILDES \citep{MacWhinney1992}

\vspace{-3mm}
\paragraph{Benchmark development in child language acquisition.} 
While there are currently only very few language acquisition benchmarks, NLP and multimodal systems bring
opportunities to ease and scale child language benchmark construction. 
For example, controlled experiments on carefully-constructed supervised benchmarks can be augmented by large video datasets of children learning a language over a long period of time.


\subsection{\includegraphics[scale=0.2]{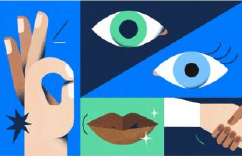} Non-Verbal Communication}

\label{sec:nonverbal}

\paragraph{Non-verbal language interpretation.} 
Non-verbal language interpretation analyzes non-verbal cues such as facial expressions, gestures, and body language to enhance the performance of NLP systems \citep{mavridis2015review, schuller2018speech}. 
For instance, while 
previous work has identified a potential ``code-book’’ of facial expressions \citep{Song2013-zg}, it remains an open research direction how to determine the  
set of expressions and gestures that can be used across modalities, contexts, and culture~\citep{matsumoto1992effects, abzaliev2022towards}. Currently, there are no LLMs that combine the gesture modality with the text.
\vspace{-3mm}
\paragraph{Sign language.} 
As a visual-gestural communication system, sign language has gained increasing attention in NLP due to its unique challenges and wide applications \citep{koller2016deep, koller2018deep,camgoz2020sign}. 
There are many research directions in sign language, such as data curation and evaluation addressing the high variability in manual gestures \citep{4563181, Li2020-hi}, incorporation of additional information, i.e., facial expressions, body pose, eye gaze~\citep{Cao2018OpenPoseRM, Baltruaitis2018OpenFace2F}; and sign language generation for various scenarios, such as speakers of the same sign language, across different sign languages, and with a combination of verbal and sign languages~\citep{signsurvey2022}. Current systems use separate models for translating sign language into the text \cite{lim2023sign}, which is then provided to LLMs. Directly providing sign language to LLM might be more efficient.

\vspace{-3mm}
\paragraph{Joint verbal and non-verbal communication.} Ultimately, both verbal and non-verbal signals should be considered during communication. Future AI systems should be equally capable of understanding ``I don’t know'', shrugging the shoulders, or  \shrug\ . 
Representing, fusing, and interpreting these signals jointly is ultimately the long-term goal of AI-assisted communication \cite{Mavridis2014ARO}. Open research problems encompass not only the development of language models for each of these modalities but effective fusion methodologies that enable large joint models for simultaneous verbal and non-verbal communication.


\section{Responsible NLP}\label{sec:responsible}
With NLP models in more applications, it is crucial to promote responsible NLP via ethical considerations (\cref{sec:ethics}), interpretability (\cref{sec:interp}), green/efficient NLP (\cref{sec:efficient}), and careful use of NLP in online environments (\cref{sec:online}).

\subsection{\includegraphics[scale=0.2]{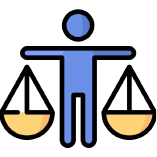} NLP and Ethics}\label{sec:ethics}

\paragraph{Dual use.} Many NLP applications that have a positive impact can at the same time be used in harmful ways \cite{Hovy2016TheSI, Wu2023UnveilingSP}. \citet{donker2023dangers} highlights an instance where LLMs were used to generate erroneous review reports, thereby disrupting the standards for peer review.
The research community should foster interdisciplinary collaboration to fight against malicious applications of NLP technologies, such as deceptive text generation, automated disinformation campaigns, LLM-powered cybersecurity attacks \cite{kang2023exploiting}, and arms racing.

\vspace{-3mm}
\paragraph{Fairness.} There is a need for methods that evaluate the fairness of NLP models, and mitigate their bias. Generative language models have been found to manifest harmful stereotypes in downstream tasks such as automated reference letter writing \cite{wan2023kelly}. While fairness in LLMs is an active area of research, much existing work focuses on limited aspects such as binary gender. Mitigating bias will involve investigating dataset creation practices and their correlation with model bias \citep{Wang2020-ve}. Such research should examine whether stricter requirements for data creation can reduce bias and inequalities that might be exacerbated by models trained on or evaluated on biased data \citep{generationGap2024}.

\vspace{-3mm}
\paragraph{Privacy.} 
With the increasing use of LLMs for personalized NLP applications, concerns have been raised regarding access to user data through LLMs, especially in sensitive areas like healthcare \citep{mesko2023imperative, marks2023ai}.
In response, researchers are investigating privacy-preserving methods such as differential privacy \citep{dwork2006differential}, federated learning \citep{fedlearning2017}, and secure multi-party computation to ensure the confidentiality and security of user data \citep{mpc2021}.


\vspace{-3mm}
\paragraph{Attribution of machine-generated data.} The use of generative LLMs in the creative industry has led to issues such as lack of copyright, plagiarism, and profit shifting.
Text generated by LLMs can reveal sensitive or copyrighted contents from
their training data.
Therefore, it is essential to develop standard approaches (e.g., membership inference \citep{membershipinf2017}) for attribution that NLP models can use while generating content 
\citep{Collins2023-nx}, especially for domains such as programming 
or creative writing \citep{Swanson2021-sy}, where LLM-generated content is on the rise.




\subsection{\includegraphics[scale=0.2]{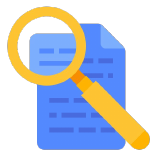} Interpretability}\label{sec:interp}

\paragraph{Probing.}
One promising direction  to investigate is the internal representation of NLP models, including LLMs, by designing probing tasks that can reveal the linguistic \cite{hewitt-manning-2019-structural,hewitt-liang-2019-designing} and world knowledge captured by the models \cite{elhage2022toy,geva-etal-2021-transformer,geva-etal-2022-transformer}. This can help understand the reasoning capabilities of models and identify potential biases \citep{li-etal-2022-quantifying, Meng2022LocatingAE}.

\vspace{-3mm}
\paragraph{Mechanistic interpretability.}
While probing mostly considers  the attributes of the features learned by the model, there are currently several open research questions around mechanistic interpretability, which  aims to uncover the underlying \textit{mechanisms and algorithms} within a model that contributes to its decision-making process \cite{nanda2023progress,conmy2023towards}. These models aim to extract computational subgraphs from neural networks \cite{conmy2023towards,wang2023interpretability,geiger2021causal}, and their high-level goal is to reverse engineer the entire deep neural network \cite{chughtai2023toy}. 
\vspace{-3mm}
\paragraph{Human-in-the-loop to improve interpretability.}
Incorporating human feedback to enhance model interpretability can
improve model transparency, facilitate better decision-making, and foster trust between AI systems and users. By involving humans, researchers can identify and address biases, ensure ethical considerations, and develop more reliable and understandable NLP models.
There are various promising directions, such as active learning and interactive explanation generation \citep{Mosca2023-vt, Mosqueira-Rey2023-cs}.
\vspace{-3mm}
\paragraph{Basing the generated text on references.}
As model-generated text is prone to hallucinations \citep{Ji2022SurveyOH},
a promising way to improve its reliability is to explain its conclusion step by step and supply references or sources to back up the claims~\citep{cot2022, Izacard2022FewshotLW}.


\subsection{\includegraphics[scale=0.2]{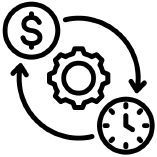} Green/Efficient NLP}\label{sec:efficient}

\paragraph{Model efficiency.} 
The trend of scaling up NLP models has accentuated the need of increasing model efficiency \citep{energy&policyStrubell2020, bridge2022}. Researchers have proposed efficiency enhancement techniques from various aspects, such as 
improving the attention mechanisms \citep{Tay2020-ti, Tay2022-zv, Dao2022-ww, Ma2022-oa}, 
sparsing models to scale up the width of models for increased expressiveness while reducing theoretical FLOPs,
and
applying mixture-of-experts architectures
\citep{Fedus2021-ol, Fedus2022-lf, Du2022-if}.
However, there are open challenges in how to develop the optimal architectures that balance economics, efficiency, and performance \citep{Mustafa2022-cw}.

\vspace{-3mm}

\paragraph{Efficient downstream task adaptation.} Increased applicability of pre-trained models requires efficient fine-tuning methods that adapt to downstream tasks~\citep{bioBERT2020,liu-etal-2023-task, bloombergGPT2023} by updating a small subset of the parameters \citep{Pfeiffer2020-kj, Moosavi2022-xd, Schick2021-uw, Hu2023LLMAdaptersAA}. For example, prompt-tuning/ prefix-tuning modifies activations with additionally learned vectors without changing model parameters \citep{Valipour2022-yt, Lester2021-ta}.

\vspace{-3mm}

\paragraph{Data efficiency.} Another method to improve efficiency is to remove redundant or noisy data in the first place.
Despite existing efforts on removing noisy examples and deduplicating data on smaller scales \citep{Lee2022-oq, Mishra2020-et, Hoffmann2022-xk}, there is a lack of effective methods for data deduplication and curation for vast corpora ($>$700B Tokens) or raw web data used for training very big models. 


\subsection{\includegraphics[scale=0.2]{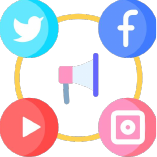} NLP for Online Environments}

\label{sec:online}

\paragraph{Combating misinformation.} 
With the rising capability of text generation models and strong commercial and political interests, it is inevitable to see an increase in online misinformation.
To fight back against powerful generative models to spread misinformation, we need to use powerful discriminative models to detect them.
There is a large need for 
fact-checking technology
\cite{thorne-etal-2018-fever, nakov2021automated, kazemi2022adaptable}, across different languages~\citep{ das2023state}, different modality ~\citep{abdelnabi2022open}, and by utilizing techniques from other areas, such as network analysis to track who likes or reposts false contents ~\citep{guarino2020beyond}, and
retrieval and knowledge-augmented methods~\citep{ciampaglia2015computational, markov2023holistic}
to search through and find the relevant context around the claim.
%
One caution is that LLMs are prone to hallucinations \cite{dziri-etal-2022-origin, raunak-etal-2021-curious} and factual inconsistencies \cite{tam-etal-2023-evaluating}, so they might not be self-sufficient to combat misinformation reliably.



\vspace{-3mm}
\paragraph{Ensuring content diversity.} With the prevalence of LLM-generated content, the majority's voice may end up amplified on the web, since data-driven models such as LLMs tend to remember the type of data that is the most represented in its corpus. Thus, the lack of diversity and especially representation of marginalized groups' voices will be a concerning problem as LLM-generated content will be increasingly used online~\cite{ Field2021ASO}.
\vspace{-3mm}
\paragraph{Preventing mis- and over-moderation.} Similar to the heterogeneity issue in content generation, content moderation techniques might also overlook the nuances of expressions in under-represented groups, or specific social environments, making them unfairly delete safe speech by minority groups \citep{sap2019risk, Xia2020DemotingRB}. 
Apart from mis-moderation, there is also over-moderation. Due to various political interests (e.g., Florida aiming to curtail discussions about race or queer identities%
), governments are likely to limit the set of topics discussed online, so it is important to trace what topics and opinions are filtered or demoted on the internet and reflect on the freedom of speech in the political environment \citep{wright2006government, doi:10.1177/2053951719897945}.






\section{Applied NLP}\label{sec:applied}
After discussing tasks in fundamental NLP and responsible NLP, we now look into the wide applications of NLP in various domains, with a few selected discussions on
NLP for healthcare (\cref{sec:health}), education (\cref{sec:edu}), computational social science (CSS) (\cref{sec:css}), and synthetic data generation (\cref{sec:synthetic}).

\subsection{\includegraphics[scale=0.2]{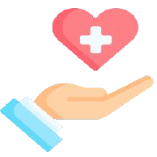} NLP for Healthcare}\label{sec:health}

\paragraph{Healthcare benchmark construction.}
Healthcare is a domain that heavily suffers from data scarcity, which is usually due to
data unavailability (typically for low-resource domains), or inaccessibility (due to privacy and ethics concerns). Potential strategies to create and scale-up health datasets include synthetic data generation \cite{chintagunta-etal-2021-medically, liednikova-etal-2020-learning, Mattern2022DifferentiallyPL} or data augmentation from existing data \citep{dai2023chataug}. These strategies 
can improve the distribution of biased datasets, help ensure data privacy protections, and reduce the cost of data collection. However, data generation by LLMs also brings concerns of
bias propagation and information leakage~\cite{Arora190}. Furthermore, researchers need metrics to measure the fidelity of synthetic data compared with real data~\cite{chen2021synthetic}.
\vspace{-3mm}
\paragraph{Improving clinical communication.}
NLP has shown great potential in enhancing communication in healthcare, such as simplifying the medical jargon for laymen \citep{jin-etal-2022-deep}, developing educational tools for healthcare professionals~\cite{min-etal-2022-pair}, and providing personalized healthcare recommendations \citep{choi2016doctor, roehrs2018toward}. New research directions
include developing advanced NLP models for medical dialogue systems and exploring the ethical implications of NLP-driven communication in healthcare \citep{ravi2020deep, holstein2019ai}. Current LLMs may only be useful in limited settings, as trust in LLMs has been shown to depend on the health-related complexity of questions \cite{info:doi/10.2196/46939}.


 \vspace{-3mm}
\paragraph{Drug discovery.} 
Since the hypothesis space for drug designs is exponential \citep{Ruddigkeit2012EnumerationO1}, NLP methods have been explored to assist clinicians to efficiently extract and analyze
information from large amounts of scientific literature, patents, 
clinical records, and other biomedical sources. Open research directions in this domain include identifying and prioritizing the drug-target interactions, discovering new drug candidates, predicting compound properties, and optimizing drug designs~\citep{Brown2020ArtificialII}. Despite their great potential,
the use of LLMs still face many challenges, such as
the lack of transparency in the model decision-making process, which limits the applicability and reliability~\cite{thirunavukarasu2023large}.  


\subsection{\includegraphics[scale=0.2]{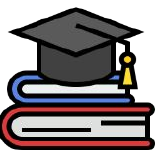} NLP for Education}\label{sec:edu}

\paragraph{Intelligent tutoring systems.} 
The rising capability of NLP systems has given rise to intelligent tutoring systems
to generate targeted practice questions and explain students’ mistakes in a wide range of subjects, from English or History to Physics or Computer Science \citep{mousavinasab2021intelligent}. 
Responsible development of these systems requires human-in-the-loop checks to ensure their reliability, 
as NLP models are still lacking when it comes to more challenging reasoning and grounding tasks~\citep{Kanda2004InteractiveRA}.
Other challenges include
lack of diverse data, both in terms of population and time, privacy concerns and trustworthiness, and the need for better evaluation mechanisms~\cite{lin2023artificial}.


\vspace{-3mm}
\paragraph{Educational explanation generation.} 
To enrich teaching materials, NLP models can also help generate explanations for complicated questions or reading materials,
as well as for automatic grading systems,
since students improve more easily when grading is justified by corresponding explanations \citep{mohler2009text}. However,
some concerns include
overreliance on the model, lack of expertise among educators \cite{redecker2017european}, 
and between real knowledge and convincingly written but unverified model output \cite{kasneci2023chatgpt}. Therefore, it is important to understand the limitations of LLMs and use them only as a tool to support and enhance learning, but not as a replacement for human teachers \cite{pavlik2023collaborating}.

\vspace{-3mm}
\paragraph{Controllable text generation.} 
In education, there is a growing need for
controllable text generation \citep{ Lee2011OnTE, zhang2022survey}. 
This is helpful, for instance, for applications aiming to introduce students to new terms by generating memorable stories corresponding to their academic skill levels, interests, 
and prior experience. However, 
it is often difficult for LLMs to ensure domain diversity of the generated text while pursuing controllability, which leads to the catastrophic forgetting problem in LLMs \cite{zhai2023investigating}. Additionally, we lack reliable evaluation techniques, as well as
dedicated benchmarks and datasets for text generation with diverse control requirements \cite{zhang2023survey}. 


\subsection{\includegraphics[scale=0.2]{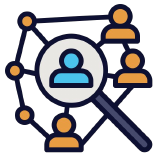} Computational Social Science}\label{sec:css}





\paragraph{Development of new abstractions, concepts, and methods.} 
NLP enables automatic analyses of massive text for the study of computational social science (CSS), which has been benefited by the evolution of NLP methods from
topic modeling \citep{blei2003latent}, keyword extraction
\citep{keyword2016}, to
word embeddings \citep{glove2014}, and LLMs \cite{brown2020gpt3}.
It is foreseeable that 
further advancement in NLP models will unlock the possibilities of more customized, high-level text analyses for CSS. Evaluation paradigms need to evolve to capture the validity of LLMs as language generators, since human evaluation also can be unreliable in CSS \cite{karpinska-etal-2021-perils}. Moreover, many CSS tasks contain large target label spaces \cite{grudin2006personas}, which is a challenge for current LLMs that have limited memory and quadratic space complexity \cite{ziems2023large}.

\vspace{-3mm}
\paragraph{Population-level data annotation and labeling.} CSS research shows a large interest in using LLMs to annotate data to simulate human interactions
\citep{Gilardi2023-pq}.
However, human studies will be unlikely to go away, 
as LLMs' effectiveness in annotation remains partial. \citet{ollion2023chatgpt} show that few-shot and zero-shot models are often outperformed by models fine-tuned with human annotations. Additionally, ChatGPT usually yields higher recall than precision, showing a tendency to output more false positives. 
}

\vspace{-3mm}
\paragraph{Multicultural and multilingual CSS.} Most CSS studies focus on English or 
other major languages, and address mostly Western cultures. However, there are many important questions in social science that require large-scale, multilingual, and multicultural analyses \citep{shen2019measuring}, such as how languages evolve, and how values vary across cultures \citep{garimella2016identifying}. This area for future work can lead to compounding impacts on the social sciences. However, the data-driven nature of LLMs makes them limited by
the under-representation of minority communities and low-resource languages in the training data.
Collecting more data related to this can help minimize the data disparity. Additionally, CSS researchers study cultures, norms, and beliefs that change across time, hence LLMs will need a high level of temporal grounding \cite{ziems2023large}.


\subsection{\includegraphics[scale=0.2]{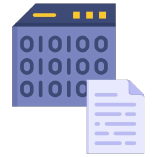} Synthetic Datasets}\label{sec:synthetic}

\paragraph{Knowledge distillation.} Knowledge distillation transfers knowledge from larger, more complex models (the teacher) to typically smaller, simpler models (the student). 
Knowledge distillation allows the knowledge and capability to be compressed into much smaller models, reducing the computational and memory requirements of NLP systems.
While earlier methods in knowledge distillation often learn from the soft output logits of teacher models \citep{Hinton2015-tk}, more recent ones 
utilize LLM outputs as synthetic examples \citep{West2022-og,Kim2022-fk}. This allows practitioners to transform or control the generated data in different ways, such as using finetuned models to filter for quality. Moreover, synthetic data can be used to directly emulate the behavior of LLMs with much smaller, focused models \citep{taori2023alpaca}. 
\vspace{-3mm}
\paragraph{Control over generated data attributes.} 
Currently, the predominant method is to provide natural text specifications with instructions and examples, but optimizing these prompts often relies on a simple trial-and-error approach. Additionally, specifying attributes through instructions or examples can be imprecise or noisy. The development of robust, controllable, and replicable pipelines for synthetic data generation remains an open research question~\citep{Kim2022-fk}.
\vspace{-3mm}
\paragraph{Transforming existing datasets.}  
Given an existing dataset, one can apply various changes to create a semantically preserving new dataset, but with a new style. Common approaches include format change (e.g., 
converting a dataset of HTML news articles  to plain text), modality transfer (e.g., generating textual descriptions of images or videos or generating captions or subtitles for audio-visual content), or style transfer~\citep{Chintagunta2021-zn, jin-etal-2022-deep} (e.g., translating the writing style of the text from verbose to concise).


\section{So ``What Should \textit{I} Work On?''}

\begin{figure}
\includegraphics[width=0.48\textwidth]{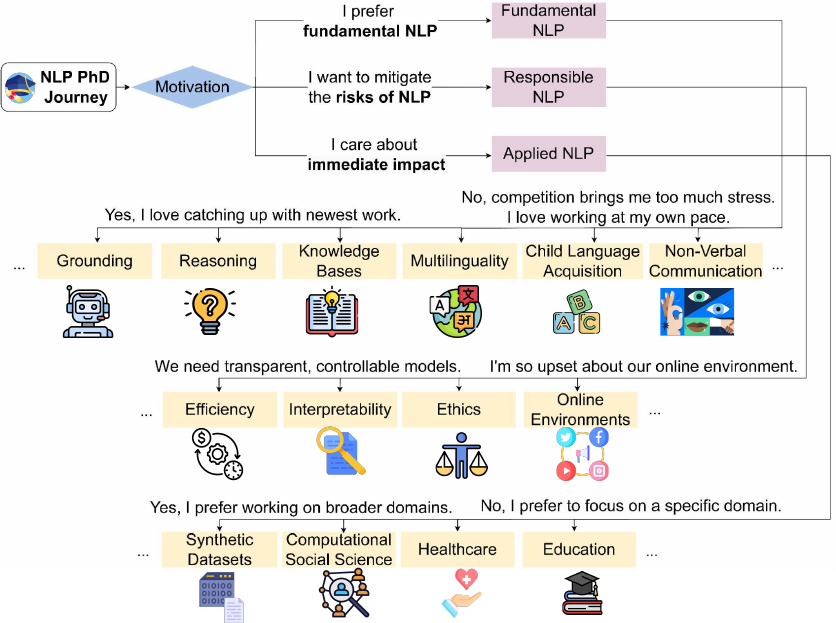}
\caption{So ``What should I work on?'' Based on your \textit{motivation}, you can choose from at least three main NLP areas: Fundamental, Responsible, and Applied. Furthermore, your choice could depend on whether or not you like fast-paced research, whether you want to make models transparent and trustworthy, or, whether you prefer broader or more specific domains. 
}
\label{fig:flow-chart}
\end{figure}

The future of NLP research is bright    
\twemoji{sparkles}. As illustrated by the 45 research directions spanning the fourteen research areas overviewed in this paper, the rapid progress we are currently witnessing in LLMs does not mean that ``it's all been solved.'' On the contrary, numerous research directions within NLP are not solved by the current LLMs. They add to the many existing tasks in NLP where LLMs' performance is limited \citep{bang2023multitask}, as well as the growing number of new areas enabled by the new LLM capabilities.

More broadly, as a field, we now have the opportunity to move away from performance-focused technology development and acknowledge that NLP is about language {\it and} people.
This brings about a new focus on enabling technologies that are culture and demographic aware, that are robust, interpretable, efficient, and aligned with solid ethical foundations ---  ultimately, technologies that make a lasting positive impact on society. 

How to choose a research direction to work on? As suggested in \Cref{fig:flow-chart}, start with your \textit{motivation} and \textit{interests}: consider your previous experiences, look around at your community, explore your curiosities about language and about people, and try to find what resonates with you the most. Building on this foundation, identify the tasks that connect to your motivations. 
This paper serves as a starting point to inspire this exploration.


\ifarxiv
\else
\section*{Broader Impact
}
We believe this work and the open research directions we identified can have an overall positive impact on the NLP research community, especially for junior students facing the  challenge of re-orienting their research directions in the era of LLMs. 

We conclude by highlighting what we foresee as the main role of this paper. First, we did not aim to cover the entire rich space of NLP, which is impossible for any research lab to enumerate exhaustively. Instead, we provided a starting point for students and researchers to regain their hope in NLP research, and find a direction they can contribute to that is not solved by LLMs.
Second, this overview paper did not aim to solve any of the tasks we listed, but rather to identify the open space for future work. We thus did not provide full details for the research directions; instead, we introduced each research direction with a brief description, its broad application, and highlight the remaining challenges and open questions, especially those that are not addressed by LLMs. Our main goal is to inspire future researchers to deepen their exploration on the topics.

We welcome suggestions for other research areas or directions to include: \url{https://bit.ly/nlp-era-llm}.
\fi


\section*{Acknowledgments}
We want to thank Steve Abney, Rui Zhang, Emily Mower-Provost, and Louis-Philippe Morency for providing feedback and valuable suggestions on earlier versions of this manuscript.
Zhijing Jin was supported by PhD fellowships from the Future of Life Institute and Open Philanthropy. This work was partially funded by 
a National Science Foundation award (\#2306372). Any opinions, findings, and conclusions or recommendations expressed in this material are those of the authors and do not necessarily reflect the views of the NSF.

\nocite{*}
\section{Bibliographical References}\label{sec:reference}

\bibliographystyle{lrec-coling2024-natbib}
\bibliography{lrec-coling2024-example}

\appendix

\section{Appendix}\label{appendix}
\appendix
\begin{table*}[!h]
\begin{tabular}{p{6cm}|p{9cm}} 
 Our Area & Standard ACL 2018 Tracks \\
\midrule
Sec 2.1 Multilinguality&	Multilinguality; Machine Translation\\
Sec 2.2 Reasoning&	Question Answering; Textual Inference\\
Sec 2.3 Knowledge Bases&	Information Extraction; Document Analysis\\
Sec 2.4 Language Grounding&	Vision, Robotics, Multimodal, Grounding and Speech\\
Sec 2.5 Child Language Acquisition&	Linguistic Theories, Cognitive Modeling and Psycholinguistics\\
Sec 2.6 Non-Verbal Communication&	-\\
Sec 3.1 NLP and Ethics&	-\\
Sec 3.2 Interpretability&	Machine Learning; Resources and Evaluation\\
Sec 3.3 Green/Efficient NLP&	-\\
Sec 3.4 NLP for Online Environments&	Social Media; Dialogue and Interactive Systems; Sentiment Analysis and Argument Mining\\
Sec 4.1 NLP for Healthcare&	Multidisciplinary; Dialogue and Interactive Systems; Information Extraction; Generation\\
Sec 4.2 NLP for Education&	Multidisciplinary; Dialogue and Interactive Systems; Information Extraction; Generation\\
Sec 4.3 Computational Social Science&	Multidisciplinary; Multilinguality; Dialogue and Interactive Systems; Sentiment Analysis and Argument Mining\\
Sec 4.4 Synthetic Datasets&	Resources and Evaluation; Generation\\
\bottomrule
\end{tabular}
\caption{
The mapping of our research areas and the ACL 2018 tracks. We address 15/ 21 areas from ACL 2018 list of research areas. The unaddressed domains, are within the purview of LLMs or outside our area of expertise (tagging and parsing): Discourse and Pragmatics; Phonology, Morphology and Word Segmentation; Sentence-level Semantics; Summarization; Tagging, Chunking, Syntax and Parsing; and Word-level Semantics.}
\label{tab:area_map}
\end{table*}
\renewcommand{\arraystretch}{1}

\end{document}